# Multidimensional classification of hippocampal shape features discriminates Alzheimer's disease and mild cognitive impairment from normal aging

Emilie Gerardin[1,2,3], MSc; Gaël Chételat[4], PhD; Marie Chupin[1,2,3], PhD; Rémi Cuingnet[1,2,3], MSc; Béatrice Desgranges[4], PhD ; Ho-Sung Kim[5], MSc; Marc Niethammer[6], PhD;  Bruno Dubois[1,2,3,7], MD, PhD ; Stéphane Lehéricy[1,2,3,8], MD, PhD; Line Garnero[1,2,3], PhD; Francis Eustache[4], PhD ; Olivier Colliot[1,2,3], PhD and the Alzheimer's Disease Neuroimaging Initiative[*]

[1]*UPMC Univ Paris 06, UMR 7225, UMR_S 975, Centre de Recherche de l'Institut Cerveau-Moelle (CRICM), Paris, F-75013, France*
[2]*CNRS, UMR 7225, CRICM, Paris, F-75013, France*
[3]*Inserm, UMR_S 975, CRICM, Paris, F-75013, France*
[4]*Inserm-EPHE-Université de Caen Basse-Normandie, Unité U923, GIP Cyceron, CHU Côte de Nacre, Caen, France*
[5]*McConnell Brain Imaging Center, Montreal Neurological Institute, McGill University, Montreal, Canada*
[6]*Department of Computer Science and Biomedical Research Imaging Center, School of Medicine, University of North Carolina at Chapel Hill, USA*
[7] *AP-HP, Department of Neurology, Groupe hospitalier Pitié-Salpêtrière, Paris, F-75013, France*
[8] *AP-HP, Department of Neuroradiology, Groupe hospitalier Pitié-Salpêtrière, Paris, F-75013, France*

**Type of manuscript**: Original research

**Correspondence to:**
Olivier Colliot
LENA - CNRS UPR 640
Cognitive Neuroscience and Brain Imaging Laboratory
Hôpital de la Pitié-Salpêtrière
47, boulevard de l'hôpital
75651 Paris Cedex 13
France
Phone: +33 1 42 16 11 86
Fax: +33 1 45 86 25 37
e-mail: olivier.colliot@upmc.fr

---

**Abstract**

We describe a new method to automatically discriminate between patients with Alzheimer's disease (AD) or mild cognitive impairment (MCI) and elderly controls, based on multidimensional classification of hippocampal shape features. This approach uses spherical harmonics (SPHARM) coefficients to model the shape of the hippocampi, which are segmented from magnetic resonance images (MRI) using a fully automatic method that we previously developed. SPHARM coefficients are used as features in a classification procedure based on support vector machines (SVM). The most relevant features for classification are selected using a bagging strategy. We evaluate the accuracy of our method in a group of 23 patients with AD (10 males, 13 females, age ± standard-deviation (SD) = 73 ± 6 years, mini-mental score (MMS) = 24.4 ± 2.8), 23 patients with amnestic MCI (10 males, 13 females, age ± SD = 74 ± 8 years, MMS = 27.3 ± 1.4) and 25 elderly healthy controls (13 males, 12 females, age ± SD = 64 ± 8 years), using leave-one-out cross-validation. For AD vs controls, we obtain a correct classification rate of 94%, a sensitivity of 96%, and a specificity of 92%. For MCI vs controls, we obtain a classification rate of 83%, a sensitivity of 83%, and a specificity of 84%. This accuracy is superior to that of hippocampal volumetry and is comparable to recently published SVM-based whole-brain classification methods, which relied on a different strategy. This new method may become a useful tool to assist in the diagnosis of Alzheimer's disease.

# Introduction

Alzheimer's disease (AD) is the most prevalent form of dementia in the elderly and the number of affected patients is expected to double in the next 20 years (Ferri et al., 2005). Accurate diagnosis of AD can be challenging, in particular at the earlier stage. Early diagnosis of AD patients is important because it allows early treatment with cholinesterase inhibitors, which have been shown to delay institutionalization, improve or stabilize cognition and behavioral symptoms (Ritchie et al., 2004; Whitehead et al., 2004). In the past years, the early clinical signs of AD have been extensively investigated, leading to the concept of amnestic Mild Cognitive Impairment (MCI) (Dubois and Albert, 2004; Dubois et al., 2007; Petersen, 2004; Petersen et al., 2001; Winblad et al., 2004). MCI patients have cognitive deficits but are capable of independent living. MCI may be symptomatic of a transition to early Alzheimer's disease.

Until recently, the role of neuroimaging in the diagnosis of AD was mainly confined to ruling out other causes of dementia. Progress in image acquisition and analysis techniques has modified this perspective: the challenge for modern neuroimaging is to help in the diagnosis of early AD and particularly in amnestic MCI patients or prodromal AD (Dubois et al., 2007). Three-dimensional (3D) magnetic resonance imaging (MRI) with high spatial resolution allows visualization of subtle anatomical changes and thus can help in the detection of brain atrophy at the beginning of the disease. Histopathological studies have shown that the hippocampus is affected by neurofibrillary tangles and amyloid plaques in the earliest stages of AD (Braak and Braak, 1995; Delacourte et al., 1999). Many studies have thus used MRI to assess *in vivo* hippocampal atrophy in AD, using manual segmentation (Fox et al., 1996; Frisoni et al., 1999; Jack et al., 1992; Jack et al., 1998; Jack et al., 1997; Juottonen et al., 1999; Killiany et al., 2002; Laakso et al., 1996; Laakso et al., 1998; Lehéricy et al., 1994; Seab et al., 1988; Xu et al., 2000). These studies have demonstrated that hippocampal volumetry is a valuable marker of AD and can distinguish patients with AD from elderly controls with a high degree of accuracy (80% to 90%). However, in patients with MCI, the discriminative power of hippocampal volumetry is substantially lower (with reported accuracy ranging from 60% to 74%) (Convit et al., 1997; De Santi et al., 2001; Du et al., 2001; Pennanen et al., 2004; Xu et al., 2000). Moreover, manual segmentation of the hippocampus requires a high degree of anatomical training, is observer-dependent and time consuming (more than one hour).

We previously developed a fully automatic method to segment the hippocampus on MRI (Chupin et al., 2009; Chupin et al., 2007). This method has been compared to manual segmentation in young healthy participants and patients with AD and has proved to be reliable, fast and accurate (about 8% relative volume error when compared to manual segmentation). We have evaluated the accuracy of automatic hippocampal volumetry to distinguish between patients with AD, MCI and elderly controls and found that it is similar to that of manual volumetry (84% for AD vs controls discrimination, 73% for MCI vs controls) (Chupin et al., 2008; Colliot et al., 2008).

However, volumetric analysis only assesses global changes of the hippocampus. On the other hand, shape analysis methods can unveil local atrophy of the hippocampus and may thus be more sensitive than volumetry, in particular at the MCI stage. Radial mapping has been used to assess local atrophy of the hippocampus in AD and MCI (Apostolova et al., 2006a; Apostolova et al., 2006b; Frisoni et al., 2006). In concordance with histopathological studies, marked atrophy was found in a region corresponding to the CA1 subfield of the hippocampal formation and the subiculum (Frisoni et al., 2006). Other studies relying on the high-dimensional brain mapping (HDBM) method (Csernansky et al., 2000; Wang et al., 2006) and a voxel-based approach (Chételat et al., 2008) also found a predominant atrophy in CA1. However, except for (Csernansky et al., 2000; Wang et al., 2006), these studies were restricted to group analysis and, in their present form, most of these methods cannot be used to classify individual patients. Csernansky et al. (2000) used the HDBM approach to classify patients with very mild AD and controls. However, they did not investigate the classification of MCI patients and it remains unclear whether the diagnostic accuracy of this approach is superior to that of volumetry.

Recently, there has been a growing interest in the use of multidimensional classification methods, such as support vector machines (SVM) (Burges, 1998; Cristianini and Shawe-Taylor, 2000), to assist in the diagnosis of neurological and psychiatric pathologies (Fan et al., 2008; Fan et al., 2007; Kloppel et al., 2008; Lao et al., 2004; Vemuri et al., 2008). In particular, several methods have been successfully applied to the classification of patients with AD or MCI. These approaches were based on the classification of anatomical features extracted from a set of regions distributed across the whole brain. However, these methods did not include a detailed morphological analysis of the hippocampus, which is affected at the earliest stages of the pathological process and may thus also provide relevant information for the classification of patients.

In this paper, we introduce a method to automatically discriminate AD and MCI patients from healthy controls, based on hippocampal shape features. Shape features are extracted using spherical harmonics, a parametric boundary description approach which can be seen as a 3D analog of Fourier series (Gerig et al., 2001a; Kelemen et al., 1999; Styner et al., 2004). Spherical harmonics coefficients are used as features in a multidimensional classification procedure based on support vector machines.

# Materials and methods

## Participants

The regional ethics committee approved the study and written informed consent, given by the patients themselves, was obtained from all participants. We studied 23 patients with AD (10 males, 13 females, age ± standard-deviation (SD) = 73 ± 6 years, range = 62–81 years, mini-mental score (MMS) = 24.4 ± 2.8, range = 19-29) and 23 patients with amnestic MCI (10 males, 13 females, age ± SD = 74 ± 8 years, range = 55–87 years, MMS = 27.3 ± 1.4, range = 24-29) recruited at the Centre Hospitalo-Universitaire (CHU) of Caen. The diagnosis for probable AD was made according to the NINCDS-ADRDA (National Institute of Neurological and Communicative Diseases and Stroke-Alzheimer's Disease and Related Disorders Association) criteria (McKhann et al., 1984). The diagnosis of MCI was based on Petersen et al.'s criteria (Petersen et al., 2001). All MCI patients were evaluated every 6 months over an 18-month period to assess conversion, i.e., whether they met NINCDS-ADRDA criteria of probable AD or not. Patients were declared as converters if they had impaired performances (more than 1.5 SD below the normal mean according to age and education when available) in at least one of general intellectual function scales as well as in at least two areas of cognition including memory, leading to impaired daily activities as judged by the clinicians from the consultation interviews. Post hoc exclusion criteria included presence of substantial neurological, psychiatric or any other medical disease that could affect brain functioning or structure, and normal episodic memory performances at follow-up. At completion of the 18-month follow-up period, seven MCI (7/22=32%) patients were declared as converters, 15 patients still had isolated memory deficits (non-converters) and one MCI patient refused follow-up. The annual conversion rate was thus 21%. AD and MCI patients were compared to 25 elderly healthy controls (13 males, 12 females, age ± SD = 64 ± 8 years, range = 51–84 years) with normal memory performance, as assessed using tests of episodic, semantic and working memory, and without vascular lesions on MRI. To exclude vascular lesions, all controls were checked to have normal signal intensity on T1-, T2- and/or FLAIR-weighted MRI, and notably no substantial white matter T2-FLAIR-weighted hyperintensities (less than 5 pinpoint hyperintensities, size <4 mm (Meguro et al., 2000)). The controls were screened for the absence of cerebrovascular risk factors, mental disorder, substance abuse, head trauma, substantial MRI or biological abnormality, and incipient dementia using a memory test battery. Control participants were recruited through advertisement in local newspapers. Control participants were required to be over 50 years old. There was no specific sex criterion.

## MRI acquisition

Within an interval of two months at most from inclusion for the controls and a few days for MCI and AD patients, each participant underwent a T1-weighted volume MRI scan, which consisted of a set of 128 adjacent axial slices parallel to the anterior commissure-posterior commissure (AC-PC) line and with slice thickness 1.5 mm and pixel size 0.9375x0.9375mm$^2$ using the spoiled gradient echo sequence (SPGR) (repetition time (TR)=10.3 ms; echo time (TE)=2.1 ms; field of view (FOV)=24*18cm²; matrix=256*192). All the MRI data sets were acquired on the same scanner (1.5 T Signa Advantage echospeed; General Electric, Milwaukee, WI).

## Automatic hippocampal segmentation

The segmentation of the hippocampus was performed using a fully automatic method we previously developed (Chupin et al., 2009; Chupin et al., 2007; Colliot et al., 2008). This approach segments both the hippocampus and the amygdala simultaneously based on competitive region-growing between these two structures. It includes prior knowledge on the location of the hippocampus and the amygdala derived from a probabilistic atlas and on the relative positions of these structures with respect to anatomical landmarks which are automatically identified.

## SPHARM description

Each hippocampus was then described by a series of spherical harmonics, using the SPHARM-PDM (Spherical Harmonics-Point Distribution Model) software developed by the University of North Carolina and the National Alliance for Medical Imaging Computing (http://www.namic.org/Wiki/index.php/Algorithm:UNC:Shape_Analysis). SPHARM are a mathematical approach to represent surfaces with spherical topology, which can be seen as a 3D analog of Fourier series expansion.

In brief, the SPHARM approach relied on the following steps. Since the automatic segmentation is based on homotopic deformations, no topological correction was necessary. Hippocampal segmentations were then converted to surface meshes, and a spherical parameterization was then computed using the approach proposed in Brechbühler et al. (1995), creating a one-to-one mapping between each point on the surface and each point on a sphere. The surface $v(\theta,\varphi) = (x(\theta,\varphi), y(\theta,\varphi), z(\theta,\varphi))^T$ was decomposed as:

$$v(\theta,\varphi) = \sum_{l=0}^{\infty} \sum_{m=-l}^{l} c_l^m Y_l^m(\theta,\varphi)$$

where the coefficients $c_l^m$ are three-dimensional vectors due to the three coordinate functions, and $Y_l^m(\theta,\varphi)$ are spherical harmonics basis functions of degree $l$ and order $m$, with $\theta \in [0,\pi], \varphi \in [0,2\pi[$.

They are defined as: $Y_l^m(\theta,\varphi) = \sqrt{\frac{(2l+1)(l-m)!}{4\pi(l+m)!}} P_l^m(\cos\theta) e^{im\varphi}$,

where $P_l^m$ are the associated Legendre polynomials

This function family is orthonormal over both $l$ and $m$: $\frac{1}{4\pi} \int_{\theta=0}^{\pi} \int_{\varphi=0}^{2\pi} Y_l^m \overline{Y_{l'}^{m'}} d\theta d\varphi = \delta_{ll'} \delta_{mm'}$

with $\delta_{ij} = 0$ if i≠j, and $\delta_{ii} = 1$ (Kronecker delta)

The series was then truncated at a given degree (here we empirically chose a degree of $L=20$ which results in an acceptable degree of smoothing). The coefficients of the series expansion were normalized in order to eliminate effects of rotation and translation: the parameterization was rotated such that the poles of the sphere match with those of the first order ellipsoid (computed from the first three SPHARM coefficients). The SPHARM representation was transformed into a triangulated surface (called the SPHARM-PDM), based on a uniform subdivision of the spherical parameterization. Each hippocampus was described by a set of 4002 landmarks. The SPHARM-PDM were finally spatially aligned using rigid Procustes alignment. To that purpose, we created a template by averaging all hippocampal surfaces. Specifically, for each of the 4002 landmarks, we computed the arithmetical mean over the whole population. This resulted in an average hippocampal surface formed by the 4002 mean landmarks. Each individual hippocampus was then aligned with respect to that average template. A new template was then formed by averaging the aligned surfaces using the procedure described above. This process was iterated until convergence of the template (i.e. when the template was no longer modified). This alignment resulted in a one-to-one mapping between points of each hippocampus. The corresponding rigid-body transform was then applied to the SPHARM decomposition, resulting in a new set of SPHARM coefficients.

We obtained two types of correspondences between objects:
- SPHARM coefficients, which were entered as features in the SVM classification;
- SPHARM-PDM landmarks, which were used to visualize the localization of shape differences between groups (Styner et al., 2004).

**Feature Extraction and Selection**

The classification features were based on the SPHARM coefficients. Specifically, each subject was represented by two sets (one for each hippocampus) of three-dimensional SPHARM coefficients $c_l^m$. When considering a SPHARM decomposition up to degree 20, each subject can thus be represented by a feature vector of size 2646 which is obtained by concatenating the three coordinates of all coefficients for both sides. Indeed, decomposition up to degree 20 results in $(20+1)^2$ vector coefficients. Moreover, there are 2 hippocampi and 3 spatial coordinates. There are thus $2 \times 3 \times (20+1)^2 = 2646$ features. In the following, the vector corresponding to the *k*-th subject is denoted as: $x^k = (x_1^k, ..., x_n^k) \in \Re^n$ where $n=2646$ is the number of features.

Among these features, only some of them convey relevant information for the classification of patients. To identify the most discriminative features for the SVM classification, we used a univariate feature selection combined with a bagging strategy. We used Student's t-tests in order to determine which features best separate the different populations. In order to obtain a robust selection, the T statistics were computed using a bagging approach (Breiman, 1996; Fan et al., 2007). This approach proceeds as follows:

For the *i*-th feature, we computed the $T_i$ statistics as follows

- For the *k*-th subject, we computed the $T_i$ statistics from the set $X^{(k)} = X - \{x^k\}$
- In order to keep only those features which are always significantly different, we computed $T_i = \min_k T_i^{(k)}$

We repeated this procedure for each of the initial features, and sorted them by increasing $T_i$. The *p* features which result in the highest $T_i$ values were kept as features in the SVM. The selected features were then centered and normalized using a z-score. In the following, in order to simplify notations, we denote the vector of selected, normalized, sorted, and centered features as $x^k = (x_1^k, ..., x_p^k) \in \Re^p$.

**Classification using SVM**

A support vector machine is a supervised learning method. In brief: given a training set $\{x^k, y^k\}_{k=1...K}$, where $x^k \in \Re^p$ are observations, and $y^k \in \{-1, 1\}$ are corresponding labels (-1 for controls, 1 for patients for example), linear SVMs search for the optimal hyperplane separating groups, i.e. the hyperplane for which the margin between groups is maximal. To that purpose, the following constrained optimization problem is solved:

$$\min_{w,b,\xi} \frac{1}{2} w^T.w + C \sum_{k=1}^{K} \xi^k$$

subject to $y^k(w^T.x^k + b) \geq 1 - \xi^k$

$$\xi^k \geq 0$$

where C is a cost parameter and the $\xi^k$ are positive slack variables allowing some examples to lie on the wrong side of the soft margin. Let the classification function be: $f(x) = \text{sign}(x.w + b)$ where *w* determines the orientation of the hyperplane, and *b* the offset from the origin. The vector *w* maximizing the margin can be written as a linear combination of some training examples, called support vectors. The classification function depends only on dot products of the data. SVM can be used to construct nonlinear separations by replacing the dot product with a kernel evaluation into the original problem. Here, we chose to use Radial Basis Functions kernels (RBF): $K_\gamma(u,v) = e^{-\gamma \|u-v\|^2}$, where the parameter γ controls the width of the kernel. We also compared the performance of the RBF kernel with that of a linear kernel. The SVM implementation relied on the LIBSVM Library (http://www.csie.ntu.edu.tw/~cijlin/libsvm). More details on SVM can be found in Cristianini and Shawe-Taylor (2000).

**Validation**

Classification accuracy (proportion of subjects correctly classified among the whole population), error (proportion of subjects wrongly classified), sensitivity (proportion of AD or MCI patients correctly classified) and specificity (proportion of healthy controls correctly classified) were computed using leave-one-out cross validation. To avoid introducing bias in the feature selection, the feature selection step detailed above was integrated in the leave-one-out procedure. In this procedure, each subject was successively selected as the test subject and all remaining subjects were used for the feature selection and classifier training. To estimate the optimal parameters C and γ of the SVM, we used a grid search with values ranging from $C=2^0 \ldots 2^{20}$ and $\gamma=2^{-15} \ldots 2^0$. We computed the classification accuracy for different number of selected features (corresponding to different thresholds on the p-value of the T statistic).

**Comparison with a voxel-based SVM approach**

We compared our approach to that proposed by Klöppel et al. (2008) in the same context of automatic classification of patients with AD. Klöppel et al. (2008) propose two different versions of their approach: one is based on whole-brain data and the other includes only data from a region of interest (ROI) located in the anterior medial temporal lobe, including part of the hippocampus. This allows comparing the accuracy of our approach to both a whole-brain approach and an approach restricted to the hippocampal region. In brief, the approach of Klöppel et al. (2008) relied on the following steps. Images were first segmented into gray matter (GM), white matter and cerebrospinal fluid using SPM5 (Statistical Parametric Mapping, Institute of Neurology, University College London, London, UK). Then, GM segmentations were further normalized to the population templates generated from all the images involved in each classification experiment (i.e. all AD and control subjects for the AD vs controls classification experiment, all MCI and controls for MCI vs controls classification), using the DARTEL diffeomorphic registration algorithm (Ashburner et al., 2007). Tissue classes were then modulated to ensure that the overall amount remained constant. No spatial smoothing was performed. Kernel matrices were then created from normalized GM segmented images and used to classify patients using linear and RBF SVM. The classification was then performed using the two different types of analysis: the first one using whole-brain data and the second one using data from a hippocampus-centered ROI. Leave-one-out accuracies were optimized with respect to C and γ with the same range of parameters as in our method.

In our method, we selected the optimal number of features based on multiple leave-one-out experiments. This can introduce a bias in the evaluation. No such bias is present in Klöppel et al's method. In order to fairly compare these two approaches, we need to evaluate our method without the bias concerning the number of selected features. We therefore tested both approaches on a completely separate group, while computing the threshold on the p-value for feature selection on the original group. To that purpose, we randomly selected from the Alzheimer's Disease Neuroimaging Initiative (ADNI) database (www.loni.ucla.edu/ADNI) 25 AD patients (age: 75 ± 6 years, MMS=23.3±2), 25 patients with amnestic MCI (age: 75 ± 5 years, MMS= 26.6±1.8), and 26 elderly cognitively normal individuals (age: 75 ± 4 years, MMS=28.9±1.3). The ADNI was launched in 2003 by the National Institute on Aging (NIA), the National Institute of Biomedical Imaging and Bioengineering (NIBIB), the Food and Drug Administration (FDA), private pharmaceutical companies and non-profit organizations, as a $60 million, 5-year public-private partnership. The primary goal of ADNI has been to test whether serial magnetic resonance imaging (MRI), positron emission tomography (PET), other biological markers, and clinical and neuropsychological assessment can be combined to measure the progression of mild cognitive impairment (MCI) and early Alzheimer's disease (AD). Determination of sensitive and specific markers of very early AD progression is intended to aid researchers and clinicians to develop new treatments and monitor their effectiveness, as well as lessen the time and cost of clinical trials. The Principal Investigator of this initiative is Michael W. Weiner, M.D., VA Medical Center and University of California - San Francisco. ADNI is the result of efforts of many co-investigators from a broad range of academic institutions and private corporations, and subjects have been recruited from over 50 sites across the U.S. and Canada. The initial goal of ADNI was to recruit 800 adults, ages 55 to 90, to participate in the research – approximately 200 cognitively normal older individuals to be followed for 3 years, 400 people with MCI to be followed for 3 years, and 200 people with early AD to be followed for 2 years. For up-to-date information see www.adni-info.org.

The complete classification procedure was the same as before, apart from feature selection.

For feature selection, instead of optimizing the number of features, we used the optimal p-value computed for the original group (p=0.002) and applied it directly to the ADNI group. There is no bias in this selection since no data from the ADNI group was used to compute the p-value threshold. We also tested the performance obtained on the original group when the optimal p-value is computed on the ADNI group.

**Comparison with classification approaches based on SPHARM-PDM**
We compared the accuracy of classification based on SPHARM to that of classification based on the SPHARM-PDM. To that purpose, we performed two experiments. In the first experiment, we applied to the SPHARM-PDM the same univariate feature selection strategy that we used in our method for the SPHARM coefficients. We computed the classification accuracies obtained with a SVM, for varying numbers of features. In the second experiment, the dimensionality of the SPHARM-PDM was reduced using principal component analysis (PCA). The resulting eigenvectors were ranked by decreasing order according to their corresponding eigenvalues. We computed the classification accuracies obtained with a SVM, for varying numbers of eigenvectors.

We also compared our approach to that proposed by Shen et al. (2003a). This approach relied on the following steps: each hippocampus was described by a set of SPHARM-PDM landmarks (the number of landmarks was 642 as described in (Shen et al., 2003a)). The dimensionality of the SPHARM-PDM was reduced using PCA. The eigenvectors were ranked in decreasing order. Fisher's Linear Discriminant (FLD) was used as a classifier. This is quite similar to the experiment performed above except that it relies on FLD instead of SVM. As indicated in (Shen et al., 2003a), two experiments (one for the left hippocampi, the other for the right) were performed, and classification was computed while letting the number of principal components (PC) vary.

**Statistical group analysis using SPHARM-PDM**
In order to illustrate the behaviour of the SPHARM approach for the detection of local hippocampal abnormalities, we used the SPHARM-PDM to perform a statistical group analysis investigating the differences between AD/MCI patients and healthy controls. It should be noted that this analysis is presented here only for illustrative purposes, in order to compare the behaviour of SPHARM with other hippocampal morphometric studies in AD, and that the SPHARM-PDM were not used for the classification of subjects.

To test for group differences in the spatial location ($x,y,z$) at each vertex of the hippocampal surface, we used the multivariate Hotelling $T^2$ metric (Styner et al., 2006). For each group, the mean $\mu_i$ and the covariance matrix $\Sigma_i$ of the spatial location were computed. Then the modified $T^2$ metric at each vertex was given by:

$$T^2 = (\mu_1 - \mu_2)^T (\frac{1}{K_1}\Sigma_1 + \frac{1}{K_2}\Sigma_2)^{-1} (\mu_1 - \mu_2)$$

where $K_i$ is the number of subjects of the *i*-th group.

P-values were obtained via permutation tests. This analysis was repeated at each vertex, resulting in a significance map, which shows local group differences. The p-values were corrected for multiple comparisons using a permutation-based approach (Styner et al., 2006).

# Results

## Classification results

The classification accuracy for a varying number of features is shown on Figures 1 and 2. For AD vs controls classification, the best classification was obtained for a number of features between 16 and 22 (which corresponds to a p-value of 0.002), giving a correct classification rate of 94%, a sensitivity of 96%, and a specificity of 92%. For MCI vs controls, the best classification was obtained for a number of features between 2 and 3 (which also corresponds to a p-value of 0.002), giving a correct classification rate of 83%, a sensitivity of 83%, and a specificity of 84%. The accuracy of the RBF classifier was always superior (or equal) to that of the linear classifier. Figures 3 and 4 show the classification accuracy for different values of the parameters C and $\gamma$, for an optimal number of features (i.e. 19 for AD, 3 for MCI). There is a substantially large set of values which lead to optimal accuracy.

To assess the influence of the degree of decomposition on the classification performance, we repeated the selection and classification procedure with different values of maximal degree L (L=5, L=10, L=15, L=20). For AD vs controls classification, the accuracy was 92% for L=5 and L=10, and 94% for L=15 and L=20. For MCI vs controls classification, the accuracy was 83% for L=5, 10, 15 and 20.

SPHARM coefficients cannot be used directly to visualize the localization of shape changes because SPHARM basis functions have a global support across the sphere. Thus, coefficients are not associated to a localized area, but rather to a mode of deformation. To illustrate the influence of discriminative coefficients on hippocampal shape, we created a mean shape by averaging all coefficients across the AD and controls group. We chose the two most discriminative coefficients for the AD vs controls separation. We then illustrated the modes of variation by adding or subtracting 2SD of these two most discriminative coefficients (Figure 5).

## Comparison with a voxel-based SVM approach

The comparison between our approach and the one proposed by Klöppel et al. (2008) is presented in Table 1. On the ADNI group, our method reached 88% accuracy, 84% sensitivity and 92% specificity for AD vs controls and 80% accuracy, 80% sensitivity and 80% specificity for MCI vs controls. For AD vs controls classification, the accuracy reached by Klöppel's method (90%) was similar to that of our approach. For MCI vs controls, the accuracy was 71% which is substantially lower than our results.

When the optimal p-value was computed on the ADNI group and applied to the original group, our method reached 88% accuracy, 91% sensitivity and 84% specificity for AD vs controls and 79% accuracy, 78% sensitivity and 80% specificity for MCI vs controls. Thus, the results obtained when selecting the p-value on the ADNI group or on the original group are very similar.

## Comparison with classification approaches based on SPHARM-PDM

Using the SPHARM-PDM, univariate feature selection and SVM classification, we obtained the following results. AD vs controls classification reached 92% which is similar to our approach. For MCI vs controls, the accuracy reached 75% which is substantially lower than that obtained with our approach based on SPHARM coefficients (83%). Using the SPHARM-PDM, PCA and SVM classification, we obtained the following results. For AD vs controls, the best result was obtained for 15 eigenvectors, giving a correct classification rate of 94% which is the same as for the SPHARM coefficients. For MCI vs controls, the best result was obtained for five features, giving a correct classification rate of 69% which is substantially lower than with our approach.

For AD vs controls, the method proposed by Shen et al., (2003a) achieved 92% accuracy for the left hippocampi (using between 9 and 11 PC) and 96% for the right hippocampi (using 24 PC). For MCI vs controls, accuracy was 67% for the left hippocampi (with 11 PC) and 73% for the right hippocampi (8 PC), which is lower than with our approach.

Therefore, using SPHARM coefficients seems more efficient than using the PDM to discriminate between MCI patients and controls.

**Statistical group analysis using SPHARM-PDM**

The maps of group differences between AD/MCI patients and controls are shown in Figures 6 and 7. For the AD vs controls comparison, the most significant deformations are shown in the medial part of the head of the hippocampus, and in a region approximately corresponding to CA1 subfield. CA2 and CA3 regions are relatively spared. Similar but less extended patterns are found for the MCI vs controls comparison.

# Discussion

In this paper, we proposed a new approach to automatically discriminate patients with Alzheimer's disease and MCI from normal aging, based on multidimensional classification of hippocampal shape features. The SVM-based classification resulted in a high degree of accuracy.

Hippocampal shape features were extracted by expanding hippocampal surfaces into series of spherical harmonics (SPHARM). SPHARM have been applied to study morphological abnormalities of brain structures in several pathologies. However, it has been mainly used to assess group differences (Gerig et al., 2001a; Shi et al., 2007; Styner et al., 2004) rather than to assist individual diagnosis. Gerig et al. (2001b) individually classified patients with schizophrenia using the mean square distance (MSD) derived from SPHARM. However, using the MSD leads to a univariate classification in which the relationships between coefficients are not taken into account. Shen et al. (2003b) used the SPHARM-PDM landmarks as features to classify patients with schizophrenia. Here, we relied on SPHARM coefficients (and not on the SPHARM-PDM) to extract features which were used in a multidimensional SVM-based classification method. These coefficients are well suited to building classification features because they provide a multi-scale representation and thus the different features correspond to different levels of detail. In particular, the first features concentrate a lot of geometrical information. This property may be related to the fact that a relatively small number of features was sufficient to discriminate between subjects. On the other hand, it is likely that each isolated SPHARM-PDM landmark does not convey a lot of information and that a larger number of features would be necessary for the discrimination. To robustly select relevant features among the whole set of coefficients, we used a bagging strategy and leave-one-out cross-validation was performed to evaluate accuracy.

Our approach achieved classification accuracies of 94% for AD vs controls, 83% for MCI vs controls. These rates are higher to that reported for manual hippocampal volumetry in various studies, in particular for the MCI patients. Indeed, while manual hippocampal volumetry can discriminate AD patients from controls with a high degree of accuracy, ranging from 82% to 90% (Frisoni et al., 1999; Jack et al., 1992; Laakso et al., 1998; Lehéricy et al., 1994; Xu et al., 2000), the accuracy obtained for MCI patients is usually much lower, ranging from 60% to 74% (Convit et al., 1997; De Santi et al., 2001; Du et al., 2001; Pennanen et al., 2004; Xu et al., 2000). Thus, taking into account shape deformations seems to result in a higher discriminative power than considering only volume, which is a global marker of atrophy. Moreover, most volumetric studies previously relied on manual segmentation, which is time-consuming and requires specific training and is thus not suitable to clinical practice. To overcome this difficulty, we segmented the hippocampi with recently developed fully automatic software. Thus, the procedure presented in this paper, from hippocampal segmentation to classification, is fully automatic and does not require user intervention.

The optimal classification was obtained for a number of features comprised between 16 and 22 for AD patients and between 2 and 3 for MCI patients. Interestingly, these numbers of features correspond to approximately the same statistical threshold (p=0.002). This implies that, at a given statistical threshold, there are more discriminative features for AD patients than for MCI patients, which reflects the fact that atrophy is greater in AD than MCI. Thus, there are more useful features for the SVM in the case of AD patients than for MCI. When irrelevant features are added, the performance of the classifier drops. Besides, the accuracy of the RBF classifier was always superior (or equal) to that of the linear SVM. This suggests that the best separation of our data is nonlinear. Moreover, the fact that the RBF was always at least as good as the linear SVM may be related to the asymptotic behaviour of RBF. Indeed, as shown by Keerthi and Lin (2003), when $\gamma$ is close to zero and $C=\hat{C}/2\gamma$ where $\hat{C}$ is fixed, then the RBF classifier converges to a linear SVM with penalty parameter $\hat{C}$. As a result, when a complete search for parameters $\gamma$ and $C$ has been performed (which was the case in our study), the RBF classifier is at least as accurate as the linear SVM (Keerthi and Lin, 2003). Furthermore, there is a substantially large range of optimal values for parameters $C$ and $\gamma$ of the SVM. This suggests that the approach is relatively robust to the choice of these parameters. However, leave-one-out cross-validation accuracies can be optimistic, since the search for optimal parameters, as well as the selection of the number of features, occurred outside of the leave-one-out loop. As a consequence, the information about each left-out subject could help in the selection of optimal parameters C and γ, and the number of features. To be fully representative of the generalization performance, the accuracy should be computed using a three-way split. However, the relatively small sample size in our study forbids us from conducting a three-way split procedure. This is also the case in Fan et al. (2007) which reports leave-one-out accuracies. On the contrary, in Vemuri et al. (2008), the test data was completely isolated from the data that was used for parameter selection.

The accuracy was not heavily influenced by the degree of the SPHARM decomposition. For AD vs controls, the results were identical for degrees 15 and 20 and very similar for degrees 5 and 10. For MCI vs controls, the results were the same for degrees 5, 10, 15 and 20. Indeed, the selected coefficients had a degree inferior or equal to 4. Nevertheless, even if for MCI lower degrees led to a good accuracy, it still seems preferable to use a decomposition of a relatively high degree (probably around 15) in order to ensure an accurate registration of shapes.

Although lesions start in the medial temporal lobe, atrophy is also present in a distributed pattern of brain regions, including the temporal neocortex, the cingulate gyrus, the precuneus, the temporo-parietal association and the perisylvian association (Baron et al., 2001; Chételat et al., 2002; Chételat et al., 2005; Karas et al., 2003; Karas et al., 2004; Lerch et al., 2005) . Several authors have thus recently proposed to classify patients based on whole brain data and not only from a single structure (Fan et al., 2008; Kloppel et al., 2008; Vemuri et al., 2008). In these approaches, the SVM uses structural features which are extracted from a set of anatomical regions distributed across the brain. The reported classification accuracies range from 89% to 96% (Fan et al., 2008; Kloppel et al., 2008; Vemuri et al., 2008) for AD vs controls discrimination, and from 82% to 90% for MCI vs controls (Davatzikos et al., 2008a; Fan et al., 2008). Using a different strategy in which SVM features are formed from the hippocampus only, we achieved similar classification rates (94% for AD patients, 83% for MCI patients). Our approach, based on a detailed shape analysis but restricted to a single structure, and these whole brain methods, which use extensive but less detailed information, seem complementary. Ultimately, it seems interesting to combine these approaches in an integrated classification method. The AD cases included in our study were less severely affected (mean MMS=24) than those studied in Kloppel et al. (2008) (mean MMS=17) and in Vemuri et al. (2008) (median MMS=20). It should be noted that the patients studied by Kloppel et al. (2008) were pathologically confirmed. The disease severity of our AD and MCI patients was similar to those included in Fan et al. (2008) (mean MMS=23 for AD and 27 for MCI) and in Davatzikos et al. (2008a) (mean MMS=26 for MCI).

We compared our approach to that proposed by Klöppel et al. (2008). Klöppel et al. (2008) propose two different versions of their approach: one is based on whole-brain data and the other includes only data from a region of interest (ROI) located in the anterior medial temporal lobe, including part of the hippocampus. In order to perform a fair comparison between our approach and theirs, we determined the optimal threshold for feature selection on the original group of subjects and used a completely separated group for the evaluation. For MCI vs controls classification, the accuracies obtained with their method were lower than with our approach (63% for whole-brain, 71% for hippocampal ROI, compared to 80% for our method). For AD vs controls, the accuracy of their whole-brain method increased to 90%, which is slightly superior to our results (88%). This comparison experiment should be confirmed in larger groups of patients. However, these results indicate that a specific analysis of hippocampal shape may be more sensitive than an approach using a ROI, defined at the group level, to discriminate MCI patients from controls. Indeed, using a specific segmentation method dedicated to the hippocampus allows precisely defining the boundaries of this structure in each individual subject. On the contrary, Klöppel et al. (2008) use a region of interest defined at the group level which also encompasses other structures. Nevertheless, whole-brain methods, such as the one proposed by Klöppel et al. (2008), are likely to be more efficient to discriminate between different types of dementia. In particular, Klöppel et al. (2008) and Davatzikos et al. (2008b) successfully applied whole-brain approaches to the differentiation of patients with AD and FTLD (fronto-temporal lobar degeneration).

We also compared our approach based on SPHARM coefficients to classifications based on SPHARM-PDM. First, we compared our approach to a classification based on SVM and SPHARM-PDM. Second, we performed a comparison with the approach of Shen et al., (2003a) which uses a FLD classifier. In both cases, our approach reached a higher accuracy for MCI vs controls classification. This suggests that using the coefficients is more efficient than using the PDM to discriminate between MCI patients and controls.

Although ADNI MCI patients are slightly more impaired than ours (mean MMSE= 26.6 vs 27.3, respectively), this difference was not statistically significant (p=0.13), and the classification results obtained on our data were very close to those obtained on ADNI data.

Most studies of hippocampal shape in AD were restricted to group analysis and not to the discrimination of individual patients (Apostolova et al., 2006a; Apostolova et al., 2006b; Chételat et al., 2008; Csernansky et al., 2000; Frisoni et al., 2006; Shen et al., 2005; Wang et al., 2006) with the exception of Csernansky et al. (2000) in which the high-dimensional brain mapping (HDBM) approach was used to obtain hippocampal volumes and hippocampal shape differences between patients with very mild AD and controls. Using a classification based on both volume and shape features, they achieved a sensitivity of 83% and a specificity of 78%, which is lower than those obtained in our study. Moreover, they did not investigate the discrimination of MCI patients from controls. Another exception is Li et al., (2007) who proposed a method for shape analysis of the hippocampus based on SVM. An important difference between their approach and ours is that their features represent the average deformation of a surface patch from a mean surface. For AD vs controls classification, their reported accuracies are similar to ours (between 84% and 94%). However, it should be noted that the AD patients included in their study are at a more severe stage than ours (MMS=19 vs 24). Moreover, they did not investigate the classification of MCI patients.

Using SPHARM-PDM, we studied the localization of hippocampal shape changes between AD or MCI patients and controls. We found that AD patients exhibit a spatial pattern of deformations that approximately corresponds to the CA1 subfield of the hippocampus. CA2 and CA3 are relatively spared. In MCI patients, a similar but less extended spatial pattern was found. Although our method cannot provide a direct mapping of hippocampal subfields since these are below the resolution of conventional MRI, this is in concordance with histopathological studies which demonstrated that neurofibrillary tangles and neuronal loss predominate in the CA1 subfield (Hyman et al., 1984; Van Hoesen and Hyman, 1990). Previous hippocampal morphometric studies in AD have relied on different types of methodologies: the high-dimensional brain mapping (HDBM) approach (Csernansky et al., 2000; Wang et al., 2006), radial mapping (Apostolova et al., 2006b; Frisoni et al., 2006) and a voxel-based method (Chételat et al., 2008). Their main finding is that atrophy is predominant in the CA1 region.

The results of our study require confirmation in larger groups of participants. However, the use of a bagging strategy brings robustness to the feature selection. Nevertheless, the relatively small size of the samples may explain why the classification accuracy diminishes when less relevant features are added, in particular for MCI patients. In order to keep the control group as large as possible, we decided not to exclude control participants based on age. As a consequence, the mean age of the controls was significantly lower (p<0.001) than those of the AD and MCI patients. However, it should be noted that misclassified controls were not always among the oldest. The misclassified controls were respectively 53 and 84 years old for AD vs controls classification, and 59, 67, 70 and 84 years for MCI vs controls classification. Nevertheless, further studies on larger age-matched groups of participants are required to confirm our results. Moreover, we could not investigate the classification of converters vs non-converters, due to the small number of converters. Indeed, MCI is a heterogeneous population and not all MCI patients have prodromal AD. Further studies on larger groups of longitudinally followed MCI patients are needed to assess whether our method can identify patients with incipient AD in an MCI population.

Using multidimensional classification of hippocampal shape features, we were able to individually classify Alzheimer's disease, MCI and control participants with a high degree of accuracy. This method may become a useful tool to assist in the diagnosis of Alzheimer's disease.

**Acknowledgements**
Data collection and sharing for this project was funded by the Alzheimer's Disease Neuroimaging Initiative (ADNI; Principal Investigator: Michael Weiner; NIH grant U01 AG024904). ADNI is funded by the National Institute on Aging, the National Institute of Biomedical Imaging and Bioengineering (NIBIB), and through generous contributions from the following: Pfizer Inc., Wyeth Research, Bristol-Myers Squibb, Eli Lilly and Company, GlaxoSmithKline, Merck & Co. Inc., AstraZeneca AB, Novartis Pharmaceuticals Corporation, Alzheimer's Association, Eisai Global Clinical Development, Elan Corporation plc, Forest Laboratories, and the Institute for the Study of Aging, with participation from the U.S. Food and Drug Administration. Industry partnerships are coordinated through the Foundation for the National Institutes of Health. The grantee organization is the Northern California Institute for Research and Education, and the study is coordinated by the Alzheimer's Disease Cooperative Study at the University of California, San Diego. ADNI data are disseminated by the Laboratory of Neuro Imaging at the University of California, Los Angeles.

# References

Apostolova LG, Dinov ID, Dutton RA, Hayashi KM, Toga AW, Cummings JL, et al. 3D comparison of hippocampal atrophy in amnestic mild cognitive impairment and Alzheimer's disease. Brain 2006a; 129: 2867-73.

Apostolova LG, Dutton RA, Dinov ID, Hayashi KM, Toga AW, Cummings JL, et al. Conversion of mild cognitive impairment to Alzheimer disease predicted by hippocampal atrophy maps. Arch Neurol 2006b; 63: 693-9.

Baron JC, Chetelat G, Desgranges B, Perchey G, Landeau B, de la Sayette V, et al. In vivo mapping of gray matter loss with voxel-based morphometry in mild Alzheimer's disease. Neuroimage 2001; 14: 298-309.

Braak H, Braak E. Staging of Alzheimer's disease-related neurofibrillary changes. Neurobiol Aging 1995; 16: 271-8; discussion 278-84.

Brechbühler C, Gerig G, Kübler O. Parameterization of closed surfaces for 3-D shape description. Comp Vision, Graphics and Image Proc 1995; 61: 154-170.

Breiman L. Bagging predictors. Machine Learning 1996; 24: 123-140.

Burges CJ. A Tutorial on Support Vector Machines for Pattern Recognition. Data Mining and Knowledge Discovery 1998; 2: 121-167.

Chételat G, Desgranges B, De La Sayette V, Viader F, Eustache F, Baron JC. Mapping gray matter loss with voxel-based morphometry in mild cognitive impairment. Neuroreport 2002; 13: 1939-43.

Chételat G, Fouquet M, Kalpouzos G, Denghien I, De la Sayette V, Viader F, et al. Three-dimensional surface mapping of hippocampal atrophy progression from MCI to AD and over normal aging as assessed using voxel-based morphometry. Neuropsychologia 2008; 46: 1721-31.

Chételat G, Landeau B, Eustache F, Mezenge F, Viader F, de la Sayette V, et al. Using voxel-based morphometry to map the structural changes associated with rapid conversion in MCI: a longitudinal MRI study. Neuroimage 2005; 27: 934-46.

Chupin M, Chételat G, Lemieux L, Dubois B, Garnero L, Benali H, et al. Fully automatic hippocampus segmentation discriminates between early Alzheimer's disease and normal aging. Proc. IEEE International Symposium on Biomedical Imaging ISBI 2008. Paris, France: IEEE Press, 2008: 97-100.

Chupin M, Hammers A, Liu RS, Colliot O, Burdett J, Bardinet E, et al. Automatic segmentation of the hippocampus and the amygdala driven by hybrid constraints: Method and validation. Neuroimage 2009.

Chupin M, Mukuna-Bantumbakulu AR, Hasboun D, Bardinet E, Baillet S, Kinkingnehun S, et al. Anatomically constrained region deformation for the automated segmentation of the hippocampus and the amygdala: Method and validation on controls and patients with Alzheimer's disease. Neuroimage 2007; 34: 996-1019.

Colliot O, Chetelat G, Chupin M, Desgranges B, Magnin B, Benali H, et al. Discrimination between Alzheimer Disease, Mild Cognitive Impairment, and Normal Aging by Using Automated Segmentation of the Hippocampus. Radiology 2008.

Convit A, De Leon MJ, Tarshish C, De Santi S, Tsui W, Rusinek H, et al. Specific hippocampal volume reductions in individuals at risk for Alzheimer's disease. Neurobiol Aging 1997; 18: 131-8.

Cristianini N, Shawe-Taylor J. An introduction to support vector machines: Cambridge University Press, 2000.

Csernansky JG, Wang L, Joshi S, Miller JP, Gado M, Kido D, et al. Early DAT is distinguished from aging by high-dimensional mapping of the hippocampus. Dementia of the Alzheimer type. Neurology 2000; 55: 1636-43.

Davatzikos C, Fan Y, Wu X, Shen D, Resnick SM. Detection of prodromal Alzheimer's disease via pattern classification of magnetic resonance imaging. Neurobiol Aging 2008a; 29: 514-23.

Davatzikos C, Resnick SM, Wu X, Parmpi P, Clark CM. Individual patient diagnosis of AD and FTD via high-dimensional pattern classification of MRI. Neuroimage 2008b; 41: 1220-7.

De Santi S, de Leon MJ, Rusinek H, Convit A, Tarshish CY, Roche A, et al. Hippocampal formation glucose metabolism and volume losses in MCI and AD. Neurobiol Aging 2001; 22: 529-39.

Delacourte A, David JP, Sergeant N, Buee L, Wattez A, Vermersch P, et al. The biochemical pathway of neurofibrillary degeneration in aging and Alzheimer's disease. Neurology 1999; 52: 1158-65.

Du AT, Schuff N, Amend D, Laakso MP, Hsu YY, Jagust WJ, et al. Magnetic resonance imaging of the entorhinal cortex and hippocampus in mild cognitive impairment and Alzheimer's disease. J Neurol Neurosurg Psychiatry 2001; 71: 441-7.

Dubois B, Albert ML. Amnestic MCI or prodromal Alzheimer's disease? Lancet Neurol 2004; 3: 246-8.

Dubois B, Feldman HH, Jacova C, Dekosky ST, Barberger-Gateau P, Cummings J, et al. Research criteria for the diagnosis of Alzheimer's disease: revising the NINCDS-ADRDA criteria. Lancet Neurol 2007; 6: 734-46.

Fan Y, Batmanghelich N, Clark CM, Davatzikos C. Spatial patterns of brain atrophy in MCI patients, identified via high-dimensional pattern classification, predict subsequent cognitive decline. Neuroimage 2008; 39: 1731-1743.

Fan Y, Shen D, Gur RC, Gur RE, Davatzikos C. COMPARE: classification of morphological patterns using adaptive regional elements. IEEE Trans Med Imaging 2007; 26: 93-105.

Ferri CP, Prince M, Brayne C, Brodaty H, Fratiglioni L, Ganguli M, et al. Global prevalence of dementia: a Delphi consensus study. Lancet 2005; 366: 2112-7.

Fox NC, Warrington EK, Freeborough PA, Hartikainen P, Kennedy AM, Stevens JM, et al. Presymptomatic hippocampal atrophy in Alzheimer's disease. A longitudinal MRI study. Brain 1996; 119 ( Pt 6): 2001-7.

Frisoni GB, Laakso MP, Beltramello A, Geroldi C, Bianchetti A, Soininen H, et al. Hippocampal and entorhinal cortex atrophy in frontotemporal dementia and Alzheimer's disease. Neurology 1999; 52: 91-100.

Frisoni GB, Sabattoli F, Lee AD, Dutton RA, Toga AW, Thompson PM. In vivo neuropathology of the hippocampal formation in AD: a radial mapping MR-based study. Neuroimage 2006; 32: 104-10.

Gerig G, Styner M, Jones D, Weinberger D, Lieberman J. Shape analysis of brain ventricles using SPHARM. Proc. Workshop on Mathematical Methods in Biomedical Analysis, Proc MMBIA 2001: IEEE Computer Society, 2001a: 171-178.

Gerig G, Styner M, Shenton ME, Lieberman J. Shape versus size: improved understanding of the morphology of brain structures. Proc. Medical Image Computing and Computer-Assisted Intervention MICCAI 2001, 2001b: 24-32.

Hyman BT, Van Hoesen GW, Damasio AR, Barnes CL. Alzheimer's disease: cell-specific pathology isolates the hippocampal formation. Science 1984; 225: 1168-70.

Jack CR, Jr., Petersen RC, O'Brien PC, Tangalos EG. MR-based hippocampal volumetry in the diagnosis of Alzheimer's disease. Neurology 1992; 42: 183-8.

Jack CR, Jr., Petersen RC, Xu Y, O'Brien PC, Smith GE, Ivnik RJ, et al. Rate of medial temporal lobe atrophy in typical aging and Alzheimer's disease. Neurology 1998; 51: 993-9.

Jack CR, Jr., Petersen RC, Xu YC, Waring SC, O'Brien PC, Tangalos EG, et al. Medial temporal atrophy on MRI in normal aging and very mild Alzheimer's disease. Neurology 1997; 49: 786-94.

Juottonen K, Laakso MP, Partanen K, Soininen H. Comparative MR analysis of the entorhinal cortex and hippocampus in diagnosing Alzheimer disease. AJNR Am J Neuroradiol 1999; 20: 139-44.

Karas GB, Burton EJ, Rombouts SA, van Schijndel RA, O'Brien JT, Scheltens P, et al. A comprehensive study of gray matter loss in patients with Alzheimer's disease using optimized voxel-based morphometry. Neuroimage 2003; 18: 895-907.

Karas GB, Scheltens P, Rombouts SA, Visser PJ, van Schijndel RA, Fox NC, et al. Global and local gray matter loss in mild cognitive impairment and Alzheimer's disease. Neuroimage 2004; 23: 708-16.
Keerthi SS, Lin CJ. Asymptotic behaviors of support vector machines with Gaussian kernel. Neural Comput 2003; 15: 1667-89.
Kelemen A, Szekely G, Gerig G. Elastic model-based segmentation of 3-D neuroradiological data sets. IEEE Trans Med Imaging 1999; 18: 828-39.
Killiany RJ, Hyman BT, Gomez-Isla T, Moss MB, Kikinis R, Jolesz F, et al. MRI measures of entorhinal cortex vs hippocampus in preclinical AD. Neurology 2002; 58: 1188-96.
Kloppel S, Stonnington CM, Chu C, Draganski B, Scahill RI, Rohrer JD, et al. Automatic classification of MR scans in Alzheimer's disease. Brain 2008; 131: 681-9.
Laakso MP, Partanen K, Riekkinen P, Lehtovirta M, Helkala EL, Hallikainen M, et al. Hippocampal volumes in Alzheimer's disease, Parkinson's disease with and without dementia, and in vascular dementia: An MRI study. Neurology 1996; 46: 678-81.
Laakso MP, Soininen H, Partanen K, Lehtovirta M, Hallikainen M, Hanninen T, et al. MRI of the hippocampus in Alzheimer's disease: sensitivity, specificity, and analysis of the incorrectly classified subjects. Neurobiol Aging 1998; 19: 23-31.
Lao Z, Shen D, Xue Z, Karacali B, Resnick SM, Davatzikos C. Morphological classification of brains via high-dimensional shape transformations and machine learning methods. Neuroimage 2004; 21: 46-57.
Lehéricy S, Baulac M, Chiras J, Pierot L, Martin N, Pillon B, et al. Amygdalohippocampal MR volume measurements in the early stages of Alzheimer disease. AJNR Am J Neuroradiol 1994; 15: 929-37.
Lerch JP, Pruessner JC, Zijdenbos A, Hampel H, Teipel SJ, Evans AC. Focal decline of cortical thickness in Alzheimer's disease identified by computational neuroanatomy. Cereb Cortex 2005; 15: 995-1001.
Li S, Shi F, Pu F, Li X, Jiang T, Xie S, et al. Hippocampal shape analysis of Alzheimer disease based on machine learning methods. AJNR Am J Neuroradiol 2007; 28: 1339-45.
McKhann G, Drachman D, Folstein M, Katzman R, Price D, Stadlan EM. Clinical diagnosis of Alzheimer's disease: report of the NINCDS-ADRDA Work Group under the auspices of Department of Health and Human Services Task Force on Alzheimer's Disease. Neurology 1984; 34: 939-44.
Meguro K, Constans JM, Courtheoux P, Theron J, Viader F, Yamadori A. Atrophy of the corpus callosum correlates with white matter lesions in patients with cerebral ischaemia. Neuroradiology 2000; 42: 413-9.
Pennanen C, Kivipelto M, Tuomainen S, Hartikainen P, Hanninen T, Laakso MP, et al. Hippocampus and entorhinal cortex in mild cognitive impairment and early AD. Neurobiol Aging 2004; 25: 303-10.
Petersen RC. Mild cognitive impairment as a diagnostic entity. J Intern Med 2004; 256: 183-94.
Petersen RC, Doody R, Kurz A, Mohs RC, Morris JC, Rabins PV, et al. Current concepts in mild cognitive impairment. Arch Neurol 2001; 58: 1985-92.
Ritchie CW, Ames D, Clayton T, Lai R. Metaanalysis of randomized trials of the efficacy and safety of donepezil, galantamine, and rivastigmine for the treatment of Alzheimer disease. Am J Geriatr Psychiatry 2004; 12: 358-69.
Seab JP, Jagust WJ, Wong ST, Roos MS, Reed BR, Budinger TF. Quantitative NMR measurements of hippocampal atrophy in Alzheimer's disease. Magn Reson Med 1988; 8: 200-8.
Shen L, Ford J, Makedon F, Saykin A. Hippocampal Shape Analysis: Surface-based Representation and Classification. SPIE Medical Imaging 2003: Conference 5032 - Image Processing. San Diego, California, USA, 2003a.
Shen L, Ford J, Makedon F, Wang Y, Steinberg T, Ye S, et al. Morphometric analysis of brain structures for improved discrimination. Proc. Medical Image Computing and Computer-Assisted Intervention, MICCAI 2003: Spring, 2003b: 513-520.

Shen L, Saykin A, McHugh T, West J, Rabin L, Wishart H, et al. Morphometric Analysis of 3D Surfaces: Application to Hippocampal Shape in Mild Cognitive Impairment. CVPRIP 2005: 6th Int. Conf. on Computer Vision, Pattern Recognition and Image Processing in conjunction with 8th Joint Conference on Information Sciences (JCIS 2005). Salt Lake City, Utah, 2005: 699-702.

Shi Y, Thompson PM, de Zubicaray GI, Rose SE, Tu Z, Dinov I, et al. Direct mapping of hippocampal surfaces with intrinsic shape context. Neuroimage 2007; 37: 792-807.

Styner M, Lieberman JA, Pantazis D, Gerig G. Boundary and medial shape analysis of the hippocampus in schizophrenia. Med Image Anal 2004; 8: 197-203.

Styner M, Oguz I, Xu S, Brechbuehler C, Pantazis D, Lewitt J, et al. Framework for the Statistical Shape Analysis of Brain Structures using SPHARM-PDM. Open Science Workshop at MICCAI 2006. Copenhaguen, Denmark, 2006.

Van Hoesen GW, Hyman BT. Hippocampal formation: anatomy and the patterns of pathology in Alzheimer's disease. Prog Brain Res 1990; 83: 445-57.

Vemuri P, Gunter JL, Senjem ML, Whitwell JL, Kantarci K, Knopman DS, et al. Alzheimer's disease diagnosis in individual subjects using structural MR images: validation studies. Neuroimage 2008; 39: 1186-97.

Wang L, Miller JP, Gado MH, McKeel DW, Rothermich M, Miller MI, et al. Abnormalities of hippocampal surface structure in very mild dementia of the Alzheimer type. Neuroimage 2006; 30: 52-60.

Whitehead A, Perdomo C, Pratt RD, Birks J, Wilcock GK, Evans JG. Donepezil for the symptomatic treatment of patients with mild to moderate Alzheimer's disease: a meta-analysis of individual patient data from randomised controlled trials. Int J Geriatr Psychiatry 2004; 19: 624-33.

Winblad B, Palmer K, Kivipelto M, Jelic V, Fratiglioni L, Wahlund LO, et al. Mild cognitive impairment--beyond controversies, towards a consensus: report of the International Working Group on Mild Cognitive Impairment. J Intern Med 2004; 256: 240-6.

Xu Y, Jack CR, Jr., O'Brien PC, Kokmen E, Smith GE, Ivnik RJ, et al. Usefulness of MRI measures of entorhinal cortex versus hippocampus in AD. Neurology 2000; 54: 1760-7.

**Tables**

**Table 1**. Comparison between our approach and the method proposed by Klöppel et al., (2008), on the ADNI group.

|  | Our method | | Klöppel's whole brain | | Klöppel's ROI | |
|---|---|---|---|---|---|---|
|  | **Linear kernel** | **RBF kernel** | **Linear kernel** | **RBF kernel** | **Linear kernel** | **RBF kernel** |
| **AD vs. Controls** | 86% | 88% | 90% | 90% | 84% | 86% |
| **MCI vs. Controls** | 78% | 80% | 63% | 63% | 71% | 71% |

# Figures

**Figure 1**. AD vs controls classification. The classification accuracy is shown as a function of the number of selected features. Using a SVM with a RBF kernel gives the highest accuracy (94%), for a number of features between 16 and 22.

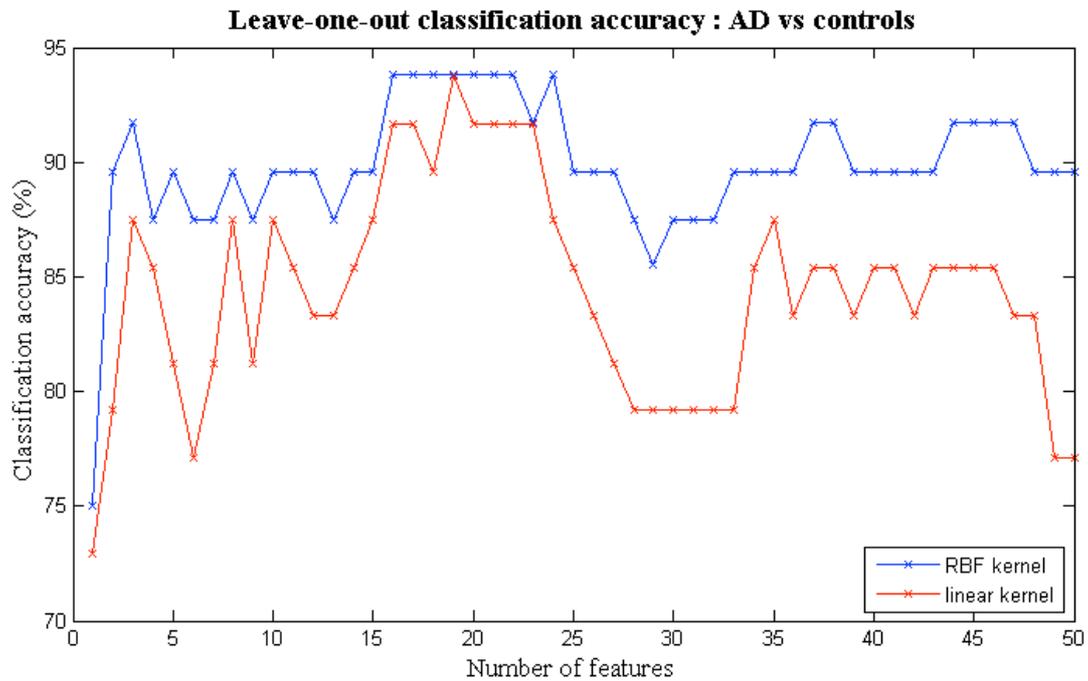

**Figure 2.** MCI vs controls classification. The classification accuracy is shown as a function of the number of features. Using a SVM with a RBF kernel gives the highest accuracy (83%), with a number of features between 2 and 3.

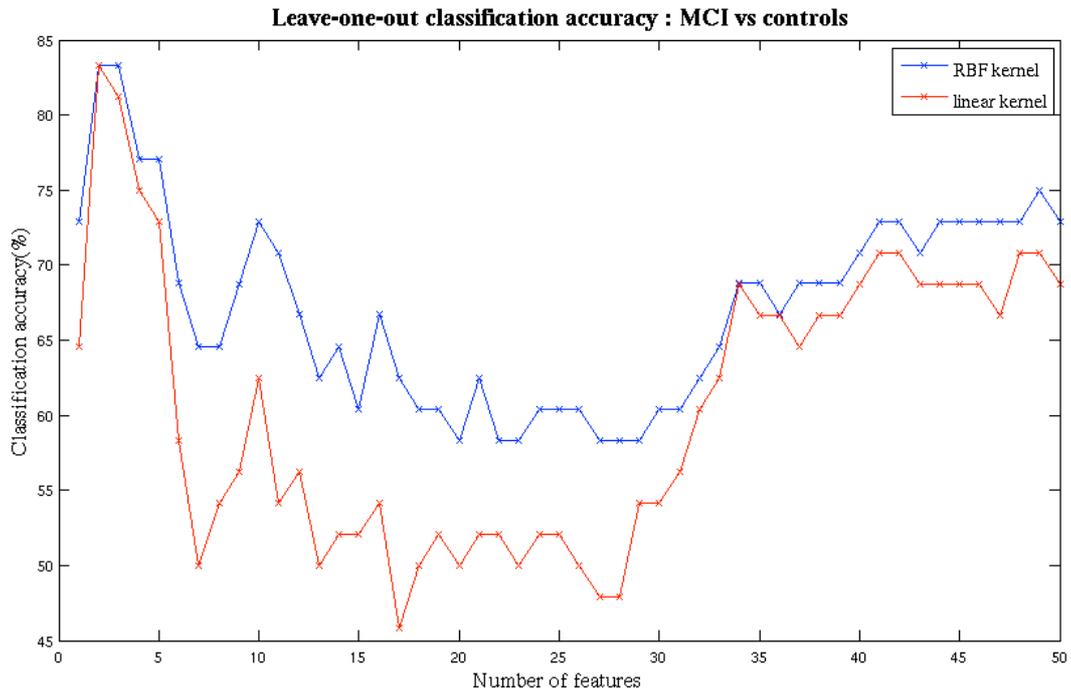

**Figure 3.** Classification rates for AD vs controls, as a function of the parameters C and γ. The following results are shown: accuracy, error, sensitivity and specificity. A "grid-search" approach was conducted in order to select the best parameters. The cost parameter C ranges from $2^0$ to $2^{20}$. The parameter γ, corresponding to the width of the RBF kernel, ranges from n $2^{-15}$ to $2^0$. Results are displayed for the classifier with 19 features, for which the best performance is obtained.

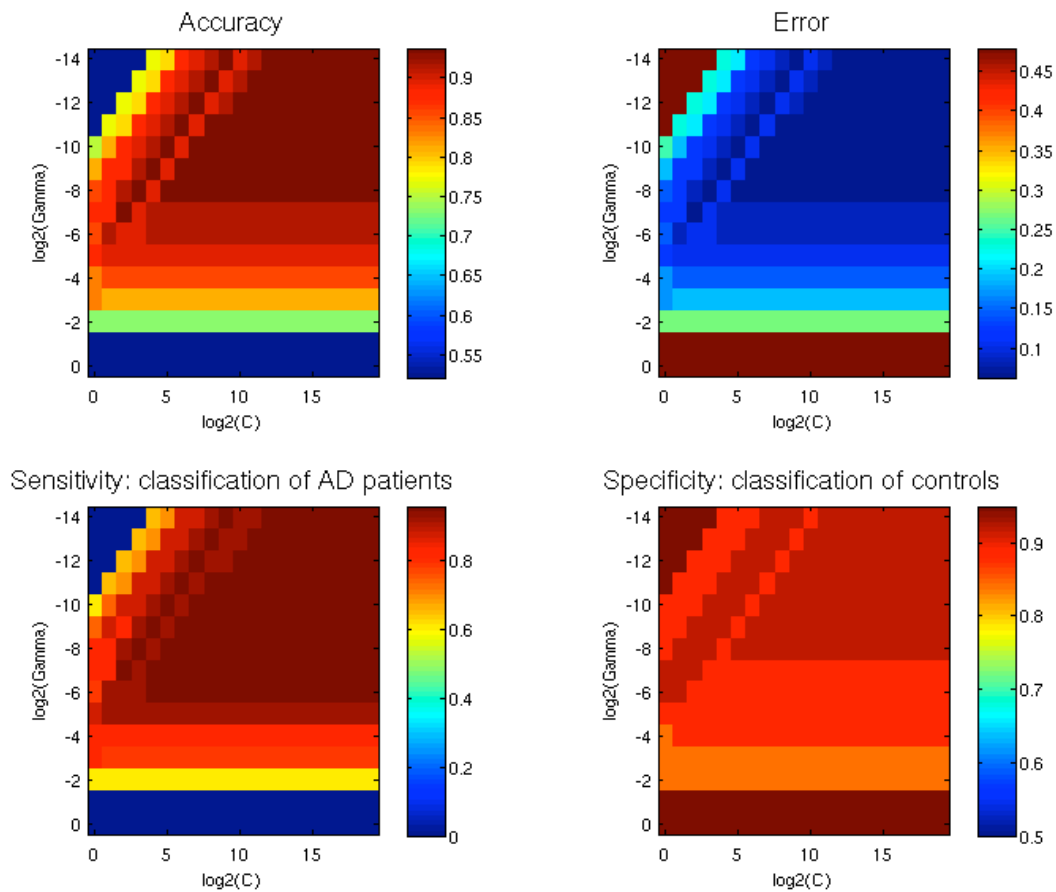

**Figure 4.** Classification rates for MCI vs controls, as a function of the parameters C and γ. The following results are shown: accuracy, error, sensitivity and specificity. A "grid-search" approach was conducted in order to select the best parameters. Results are displayed for a classifier with 3 features, for which the best performance is obtained.

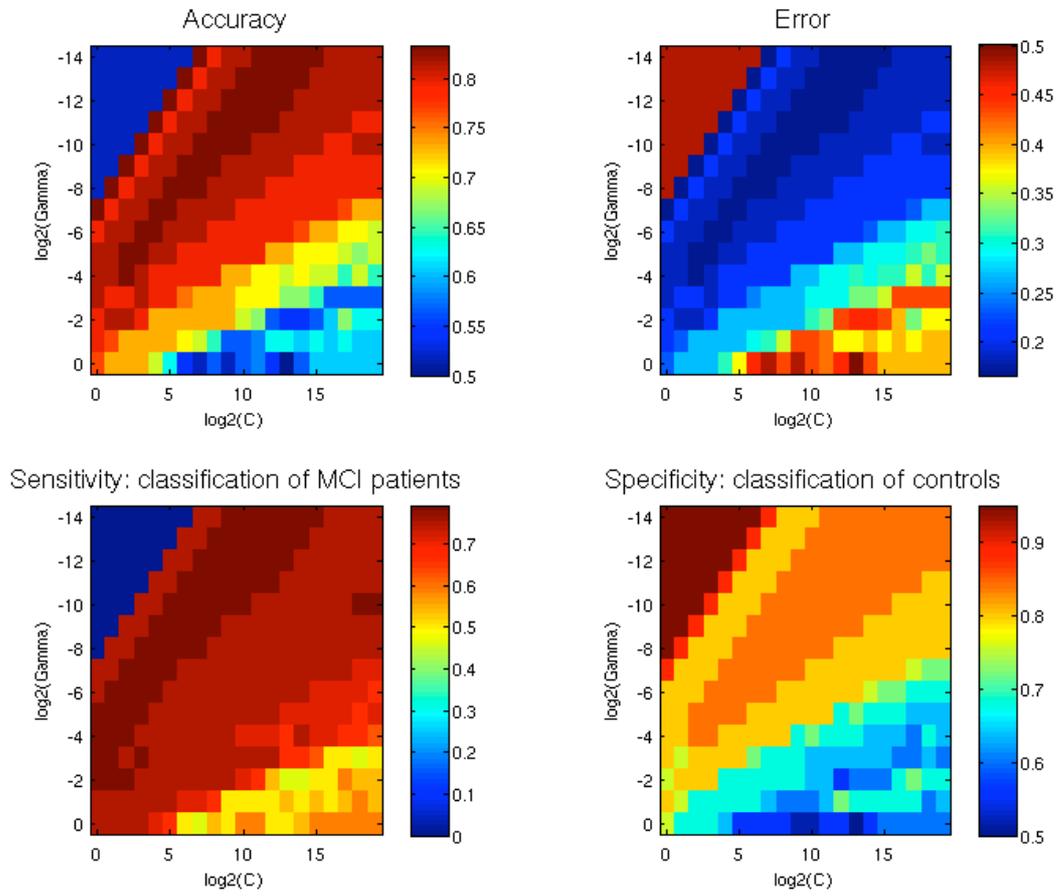

**Figure 5**. Illustration of the influence of the two most discriminative SPHARM coefficients on hippocampal shape for AD vs controls classification. A mean shape was created by averaging all coefficients across the AD and controls group. To illustrate the modes of variation, we added or subtracted to the mean shape 1SD or 2SD of the two discriminative coefficients.

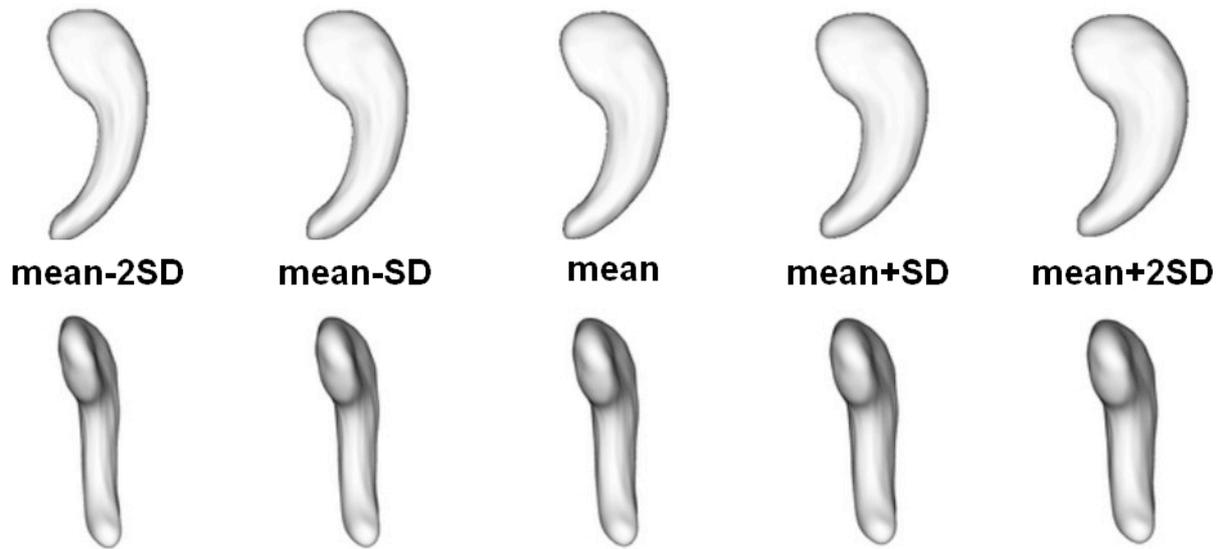

**Figure 6**. Group differences between AD and controls. Regions with statistically significant atrophy (p<0.05) are displayed in colors ranging from yellow to red, blue areas correspond to regions with no significant differences. P-values are corrected for multiple comparisons. The medial aspect of the head and the lateral aspect of the body, approximately corresponding to the CA1 subfield, are affected by deformations in AD patients.

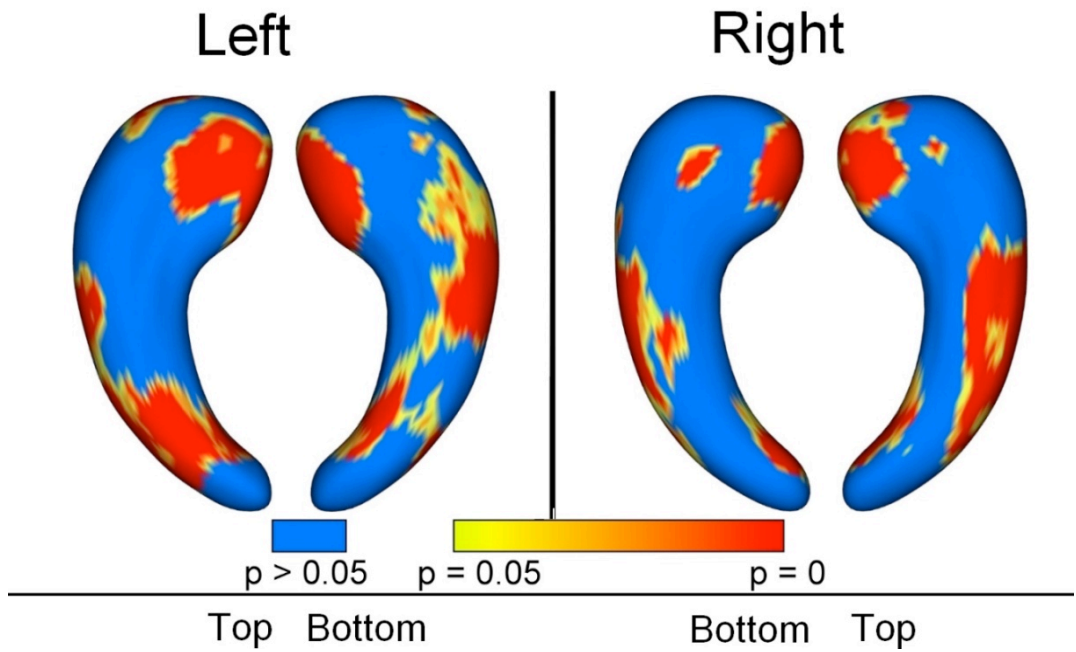

**Figure 7**. Group differences between MCI and controls. Regions with statistically significant atrophy (p<0.05) are displayed in colors ranging from yellow to red, blue areas correspond to regions with no significant differences. P-values are corrected for multiple comparisons. The patterns of atrophy are similar to those of the AD vs controls comparison but with a smaller spatial extent.

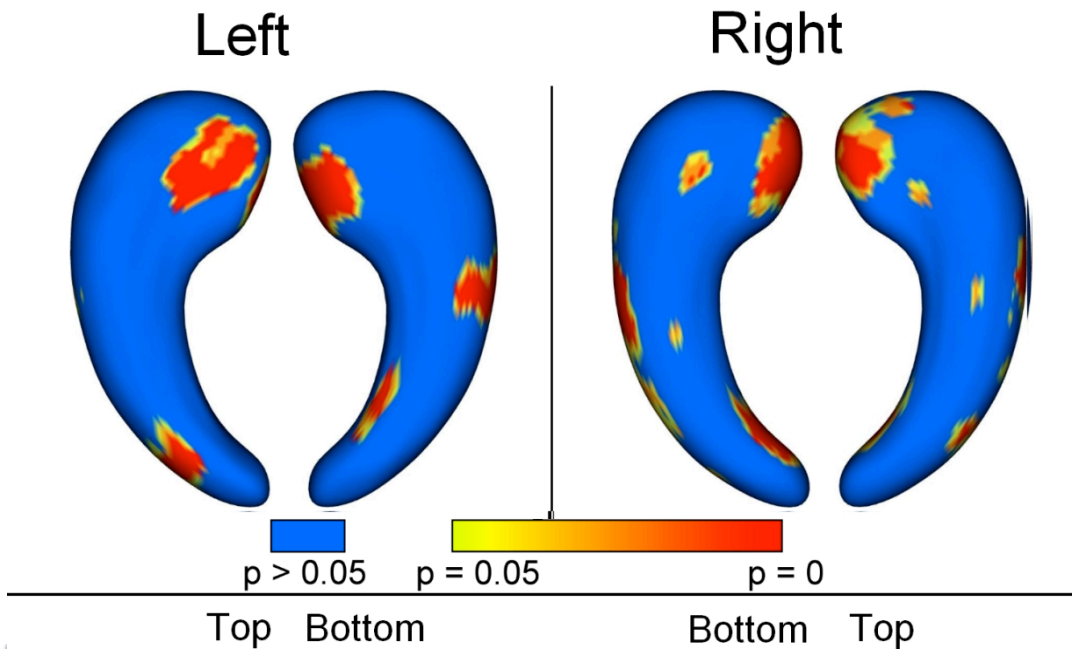